\documentclass[11pt,a4paper]{article}
\usepackage[hyperref]{emnlp2020}
\usepackage{times}
\usepackage{latexsym}
\usepackage{amsmath}
\usepackage{multirow}

\usepackage{enumitem} 
\usepackage{xfrac}

\usepackage{microtype}

\usepackage{booktabs, tabularx}
\usepackage{amssymb}
\usepackage{subfig}

\aclfinalcopy %
\def\aclpaperid{973} %

\title{
Embedding Words in Non-Vector Space\\ with Unsupervised Graph Learning
}

\author{
Max Ryabinin$^{1,2}$ \quad  Sergei Popov$^{1,2}$ \quad Liudmila Prokhorenkova$^{1,2,3}$ \quad Elena Voita$^{4,5}$\bigskip\\
  $^1$Yandex, Russia \\ $^2$National Research University Higher School of Economics, Russia\\
  $^3$Moscow Institute of Physics and Technology, Russia\\
  $^4$University of Edinburgh, Scotland \quad %
  $^5$University of Amsterdam, Netherlands \\
  {\tt \{mryab, sapopov, ostroumova-la, lena-voita\}@yandex-team.ru}
}

\date{}

\begin{document}
\maketitle
\begin{abstract}
It has become a de-facto standard to represent words as elements of a vector space~(word2vec, GloVe). While this approach is convenient, it is unnatural for language: words form a graph with a latent hierarchical structure, and this structure has to be revealed and encoded by word embeddings. We introduce GraphGlove: unsupervised graph word representations which are learned end-to-end. In our setting, each word is a node in a weighted graph and the distance between words is the shortest path distance between the corresponding nodes. We adopt a recent method learning a representation of data in the form of a differentiable weighted graph and use it to modify the GloVe training algorithm. 
We show that our graph-based representations substantially outperform vector-based methods on word similarity and analogy tasks. Our analysis reveals that the structure of the learned graphs is hierarchical and similar to that of WordNet, the geometry is highly non-trivial and contains subgraphs with different local topology.\footnote{The training algorithm, preprocessing scripts and evaluation benchmarks are available at \url{https://github.com/yandex-research/graph-glove}}
\end{abstract}

\section{Introduction}

Effective word representations are a key component of machine learning models for most natural language processing tasks. The most popular approach to represent a word is to map it to a low-dimensional vector~\cite{word2vec,pennington-etal-2014-glove,bojanowski-etal-2017-enriching,tifrea2018poincar}. Several algorithms can produce word embedding vectors with distances or dot products capturing semantic relationships between words; the vector representations can be useful for solving numerous NLP tasks such as word analogy~\cite{word2vec}, hypernymy detection~\cite{tifrea2018poincar} or serving as features for supervised learning problems.

While representing words as vectors may be convenient, it is unnatural for language: words form a graph with a hierarchical structure~\cite{miller1995wordnet} that has to be revealed and encoded by unsupervised learned word embeddings. A possible step towards this can be made by choosing a vector space more similar to the structure of the data: for example, a space with hyperbolic geometry~\cite{dhingra2018embedding,tifrea2018poincar} instead of commonly used Euclidean~\cite{word2vec,pennington-etal-2014-glove,bojanowski-etal-2017-enriching} was shown beneficial for several tasks. However, learning data structure by choosing an appropriate vector space is likely to be neither optimal nor generalizable:
\citet{gu2018learning} argue that not only are different data better modelled by different spaces, but even for the same dataset the preferable type of space may vary across its parts. It means that the quality of the representations obtained from vector-based embeddings is determined by how well the geometry of the embedding space matches the structure of the data. Therefore, 
(1)~any vector-based word embeddings inherit limitations imposed by the structure of the chosen vector space;
(2)~the vector space geometry greatly influences the properties of the learned embeddings;
(3)~these properties may be the ones of a space geometry and not the ones of a language.

In this work, we propose to embed words into a graph, which is more natural for language. In our setting, each word is a node in a weighted undirected graph and the distance between words is the shortest path distance between the corresponding nodes; note that any finite metric space can be represented in such a manner.
We adopt a recently introduced method which learns a representation of data as a weighted graph~\cite{PRODIGE}
and use it to modify the GloVe algorithm for unsupervised word embeddings~\cite{pennington-etal-2014-glove}. 
The former enables simple end-to-end training by gradient descent, the latter~--- learning a graph in an unsupervised manner. 
Using the fixed training regime of GloVe, we vary the choice of a distance: the graph distance we introduced, as well as the ones defined by vector spaces: Euclidean~\cite{pennington-etal-2014-glove} and hyperbolic~\cite{tifrea2018poincar}. This allows for a fair comparison of vector-based and graph-based approaches and analysis of limitations of vector spaces. In addition to improvements on a wide range of word similarity and analogy tasks, analysis of the structure of the learned graphs suggests that graph-based word representations can potentially be used as a tool for language analysis.

Our key contributions are as follows:
\begin{itemize}
    \item we introduce GraphGlove~--- graph word embeddings;
    \item we show that GraphGlove substantially outperforms both Euclidean and Poincar\'e GloVe on word similarity and word analogy tasks;
    \item we analyze the learned graph structure and show that GraphGlove has hierarchical, similar to WordNet, structure and highly non-trivial geometry containing subgraphs with different local topology.
\end{itemize}

\section{Graph Word Embeddings}\label{sect:method}

For a vocabulary $\mathbf{V}=\{v_0, v_1, \dots, v_n\}$, we define \textit{graph word embeddings} as an undirected weighted graph $G(V, E, w)$. In this graph, 
\begin{itemize}%
\setlength\itemsep{0em}
    \item[$\circ$] $V$ is a set of vertices corresponding to the vocabulary words;
    \item[$\circ$] $E{=}\{e_0, e_1, \dots, e_m\}$ is a set of edges: $e_i{=}e(v_{src_i}, v_{dst_i})$,\ \  $v_{src_i}$, $v_{dst_i} \ \in V$;
    \item[$\circ$] $w(e_i)$ are non-negative edge weights. 
\end{itemize}

When embedding words as vectors, the distance between words is defined as the distance between their vectors; the distance function is inherited from the chosen vector space (usually Euclidean). For graph word embeddings, the distance between words is defined as the shortest path distance between the corresponding nodes of the graph:
\begin{equation}
d_G(v_i, v_j)=\min\limits_{\pi \in \Pi_G(v_i, v_j)}\sum\limits_{e_k\in\pi}w(e_k),
\end{equation}
where $\Pi_G(v_i, v_j)$ is the set of all paths from $v_i$ to $v_j$ over the edges of $G$.

To learn graph word embeddings, we use a recently introduced method for learning a representation of data in a form of a weighted graph~\cite{PRODIGE} and modify the training procedure of GloVe~\cite{pennington-etal-2014-glove} for learning unsupervised word embeddings. We give necessary background in Section~\ref{sect:background} and introduce our method, GraphGlove, in Section~\ref{sect:graph_glove}.

\subsection{Background}
\label{sect:background}

\subsubsection{Learning Weighted Graphs}
\label{sect:background_graphs}

PRODIGE~\cite{PRODIGE} is a method for learning a representation of data in a form of a weighted graph $G(V, E, w)$. The graph requires (i) inducing a set of edges $E$ from the data and (ii) learning edge weights. To induce a set of edges, the method starts from some sufficiently large initial set of edges and, along with edge weights, learns which of the edges can be removed from the graph. Formally, it learns $G(V, E, w, p)$, where in addition to a weight $w(e_i)$, each edge $e_{i}$ has an associated Bernoulli random variable $b_i \sim Bern(p(e_i))$; this variable indicates whether an edge is present in $G$ or not. For simplicity, all random variables $b_i$ are assumed to be independent and the joint probability of all edges in the graph can be written as $p(G) = \prod_{i=0}^m p(e_i)$. Since each edge is present in the graph with some probability, the distance is reformulated as the expected shortest path distance:
\vspace{-1ex}
 $$d(v_i, v_j) =\underset{G\sim p(G)}{\mathbb{E}} d_G(v_i, v_j) =$$
 \vspace{-2ex}
\begin{equation}
\vspace{-1ex}
\begin{aligned} 
    = \underset{G\sim p(G)}{\mathbb{E}}  \ \ {\min\limits_{\pi \in \Pi_G(v_i, v_j)}} \sum_{e_i \in \pi} w(e_i),
    \label{eq:graph_distance}
\end{aligned}
\end{equation}
where $d_G(v_i, v_j)$ is computed efficiently using Dijkstra's algorithm. The probabilities $p(e_i)$ are used only in training; at test time, edges with probabilities less than $0{.}5$ are removed, and the graph~$G(V, E, w, p)$ can be treated as a deterministic graph~$G(V, E, w)$.

\paragraph{Training.} Edge probabilities $p(e_i)=p_{\theta}(e_i)$ and weights $w(e_i)=w_{\theta}(e_i)$ are learned by minimizing the following training objective:
\vspace{-1ex}
\begin{equation}
\vspace{-0.5ex}
    \mathcal{R(\theta)} = \underset{G\sim p(G)}{\mathbb{E}} \!\! \left[L(G, \theta)\right] {+} \lambda \cdot \frac{1}{|E|} \sum_{i=1}^{|E|}{p_\theta(e_i)}.
    \label{eq:objective}
\end{equation}
Here $L(G, \theta)$ is a task-specific loss, and $\frac{1}{|E|} \sum_{i=1}^{|E|}{p_\theta(e_i)}$ is the average
probability of an edge being present. The second term is the $L_0$ regularizer on the number of edges, which penalizes edges for being present in the graph. Training with such regularization results in a graph where an edge becomes either redundant (with probability close to~0) or important (with probability close to~1). 

To propagate gradients through the second term in~(\ref{eq:objective}), the authors use the log-derivative trick~\cite{glynn1990likelihood} and Monte-Carlo estimate of the resulting gradient; when sampling, they also apply a heuristic to reduce variance. For more details on the optimization procedure, we refer the reader to the original paper~\cite{PRODIGE}.

\paragraph{Initialization.} An important detail is that training starts not from the set of all possible edges for a given set of vertices, but from a chosen subset; this subset is constructed using task-specific heuristics. The authors restrict training to a subset of edges to make it feasible for large datasets: while the number of all edges in a complete graph scales quadratically to the number of vertices,
the initial subset can be constructed to scale linearly with the number of vertices.

\subsubsection{GloVe}

GloVe~\cite{pennington-etal-2014-glove} is an unsupervised method which learns word representations directly from the global corpus statistics. Each word $v_i$ in the vocabulary $V$ is associated with two vectors $w_i$ and $\tilde{w}_i$; these vectors are learned by minimizing %
\vspace{-1ex}
\begin{equation}
\vspace{-1ex}
  \mathcal{J}\!\! =\!\! \sum\limits_{i,j=1}^{|V|}\! f(X_{i,j}) (w_i^T\tilde{w}_j + b_i + \tilde{b}_j -\log X_{i,j}) ^ 2.
  \label{eq:glove_loss}
\end{equation}
Here $X_{i,j}$ is the co-occurrence between words $v_i$ and $v_j$; $b_i$ and $\tilde{b}_j$ are trainable word biases, and $f(X_{i,j})$ is a weight function: $f(X_{i,j}) = \min(1, [\frac{X_{i,j}}{x_{max}}] ^ \alpha)$ with $x_{max} = 100$ and $\alpha=3/4$.

The original GloVe learns embeddings in the Euclidean space; Poincar\'e GloVe~\cite{tifrea2018poincar} adapts this training procedure to hyperbolic vector spaces. This is done by replacing $w_i^T\tilde{w}_j$ in formula~(\ref{eq:glove_loss}) with $-h(d(w_i, \tilde{w}_j))$, where $d(w_i, \tilde{w}_j)$ is a distance in the hyperbolic space, and $h$ is either $h(d)=d^2$ or $h(d)=\cosh^2(d)$ (see Table~\ref{tab:glove_loss_overview}).

\begin{table}[t!]
\centering
\begin{tabular}{llc}
\toprule
 & & ``$\square$'' in the loss term\\
 & & $ f(X_{i,j}) (\square +\!b_i\!+\!\tilde{b}_j\!-\!\log X_{i,j})^2$\\

\midrule
\multicolumn{3}{l}{\!\!\!\!\bf Euclidean }\\
& & $w_i^T\tilde{w}_j$ \\
\midrule
\multicolumn{3}{l}{\!\!\!\!\bf Poincar\'e }\\
&\!\!\!\!$d^2$ & $-d^2(w_i, \tilde{w}_j)$ \\
&\!\!\!\!$\cosh^2 d$ & $-\cosh^2\left(d(w_i, \tilde{w}_j)\right)$ \\ 
\midrule
\multicolumn{3}{l}{\!\!\!\!\bf Graph}\\
&\!\!\!\!$d$ & $-d_G(v_i, v_j)$ \\
&\!\!\!\!\!$\langle \cdot, \cdot\rangle$ & \!\!\!\!\!\!\!\!\!  $\frac{1}{2} \left(-d_G^2(v_i, v_j)\!+\!d_G^2(v_i, 0)\!+\!d_G^2(v_j, 0)\right)$\!\!\!\\
\bottomrule
\end{tabular}
\caption{Original GloVe loss and several extensions. For Poincar\'e GloVe, $d$ is distance in the hyperbolic space; for GraphGlove, $d$ is the shortest path distance.}
\label{tab:glove_loss_overview}
\end{table}

\subsection{Our Approach: GraphGlove}
\label{sect:graph_glove}

We learn graph word embeddings within the general framework described in Section~\ref{sect:background_graphs}. Therefore, it is sufficient to (i) define a task-specific loss~$L(G, \theta)$ in formula~(\ref{eq:objective}), and (ii) specify the initial subset of edges.

\subsubsection{Loss function} 
We adopt GloVe training procedure and 
learn edge weights and probabilities directly from the co-occurrence matrix $X$. We define $L(G, \theta)$ by modifying formula~(\ref{eq:glove_loss}) for weighted graphs:
\begin{enumerate}
\setlength\itemsep{0em}
    \item replace $w_i^T\tilde{w}_j$ with either graph distance or graph dot product as shown in Table~\ref{tab:glove_loss_overview} (see details below);
    \item since we learn one representation for each word in contrast to two representations learned by GloVe, we set $\tilde{b}_j=b_j$.
\end{enumerate}

\paragraph{Distance.} We want negative distance between nodes in a graph to reflect similarity between the corresponding words; therefore, it is natural to replace $w_i^T\tilde{w}_j$ with the graph distance.
The resulting loss $L(G, \theta)$ is:
\vspace{-1ex}
\begin{equation}
  \sum_{i,j=1}^{|V|}\!\! f(X_{i,j})  ( - d_G(v_i, v_j)\! +\! b_i\! +\! b_j\!-\!\log X_{i,j}) ^ 2.
  \label{eq:graph_glove_loss_distance}
  \vspace{-1ex}
\end{equation}

\paragraph{Dot product.} A more honest approach would be replacing dot product $w_i^T\tilde{w_j}$ with a ``dot product'' on a graph. To define dot product of nodes in a graph, we first express the dot product of vectors in terms of distances and norms. Let $w_i$, $w_j$ be vectors in a Euclidean vector space, then
\begin{align}
|\!|w_i-w_j|\!|^2\!&=\!|\!|w_i|\!|^2\!+\!|\!|w_j|\!|^2\!-\!2w_i^Tw_j,\\
w_i^T\!w_j\!&=\!\frac{1}{2}\!\!\left(|\!|w_i|\!|^2\!+\! |\!|w_j|\!|^2\!\!-\!|\!|w_i\!-\!w_j|\!|^2\right)\!.
\end{align}
Now it is straightforward to define the dot product\footnote{Note that our ``dot product'' for graphs does not have properties of dot product in vector spaces; e.g., linearity by arguments.} of nodes in our weighted graph:
\begin{equation}
\vspace{-1ex}
\langle v_i, v_j\rangle \!\!=\!\! \frac{1}{2}\! \left(d^2(v_i,\! 0)\!+\!d^2(v_j,\! 0)\!-\!d^2(v_i, v_j)\right)\!,
    \label{eq:graph_dot_product}
\end{equation}
where $d(v_i, v_j)$ is the shortest path distance. 

 Note that dot product~(\ref{eq:graph_dot_product}) contains distances to a zero element; thus in addition to word nodes, we also need to add an extra ``zero'' node in a graph. This is not necessary for the distance loss~(\ref{eq:graph_glove_loss_distance}), but we add this node anyway to have a unified setting; a model can learn to use this node to build paths between other nodes.
 
 All loss functions are summarized in Table~\ref{tab:glove_loss_overview}.

\subsubsection{Initialization} We initialize the set of edges by connecting each word with its $K$ nearest neighbors and $M$ randomly sampled words. The nearest neighbors are computed as closest words in the Euclidean GloVe embedding space,\footnote{In preliminary experiments, we also used as nearest neighbors the words which have the largest pointwise mutual information (PMI) with the current one. However, such models have better loss but worse quality on downstream tasks, e.g. word similarity.} random words are sampled uniformly from the vocabulary.

We initialize biases $b_i$ from the normal distribution $\mathcal{N}(0, 0{.}01)$, edge weights by the cosine similarity between the corresponding GloVe vectors, and edge probabilities with $0{.}9$.

\section{Experimental Setup}
\label{sect:exp_setup}

\subsection{Baselines}

Our baselines are Euclidean GloVe~\cite{pennington-etal-2014-glove} and Poincar\'e GloVe~\cite{tifrea2018poincar}; for both, we use the original implementation\footnote{Euclidean GloVe: \url{https://nlp.stanford.edu/projects/glove/}, Poincar\'e GloVe: \url{https://github.com/alex-tifrea/poincare_glove}.} with recommended hyperparameters.  We chose these models to enable a comparison of our graph-based method and two different vector-based approaches within the same training scheme. %

\subsection{Corpora and Preprocessing}

We train all embeddings on Wikipedia 2017 corpus. To improve the reproducibility of our results, we (1)~use a standard publicly available Wikipedia snapshot from
\texttt{gensim-data}\footnote{\url{https://github.com/RaRe-Technologies/gensim-data} , 
dataset \texttt{wiki-english-20171001}}, 
(2)~process the data with standard GenSim Wikipedia 
tokenizer\footnote{ gensim.corpora.wikicorpus.tokenize , commit de0dcc3}. Also, we release preprocessing scripts and the resulting corpora as a part of the supplementary code.

\subsection{Setup}

We compare embeddings with the same vocabulary and number of parameters per token. For vector-based embeddings, the number of parameters equals vector dimensionality. For GraphGlove, we compute number of parameters per token as proposed by~\citet{PRODIGE}: ($|V| + 2 \cdot |E|) / |V|$. To obtain the desired number of parameters in GraphGlove, we initialize it with several times more parameters and train it with $L_0$ regularizer until enough edges are dropped (see Section \ref{sect:graph_glove}).

We consider two vocabulary sizes: 50k and 200k. For 50k vocabulary, the models are trained with either 20 or 100 parameters per token; for 200k vocabulary --- with 20 parameters per token. For initialization of GraphGlove with 20 parameters per token we set $K=64$, $M=10$; for a model with 100 parameters per token, $K=480$, $M=32$. 

In preliminary experiments, we discovered that increasing both $K$ and $M$ leads to better final representations at a cost of slower convergence; decreasing the initial graph size results in lower quality and faster training. However, starting with no random edges (i.e. $M=0$) also slows convergence down.

\subsection{Training}

Similarly to vectorial embeddings, GraphGlove learns to minimize the objective (either distance or dot product) by minibatch gradient descent. However, doing so efficiently requires a special graph-aware batching strategy. Namely, a batch has to contain only a small number of rows with potentially thousands of columns per row. This strategy takes advantage of the Dijkstra algorithm: a single run of the algorithm can find the shortest paths between a single source and multiple targets. Formally, one training step is as follows:
\begin{enumerate}[noitemsep]
    \item we choose $b=64$ unique ``anchor'' words;
    \item sample up to $n=10^4$ words that co-occur with each of $b$ ``anchors'';
    \item multiply the objective by importance sampling weights to compensate for non-uniform sampling strategy.\footnote{Let X be the co-occurrence matrix. Then for a pair of words~($v_i$, $v_j$), an importance sampling weight is $\frac{p_{i,j}}{q_{i,j}}$, where $p_{i,j}=\frac1{|\{(k, l): X_{k, l}\neq0\}|}$ is the probability to choose a pair ($v_i$, $v_j$) in the original GloVe, $q_{i,j}=\frac1{|V|}\cdot\frac1{|\{k: X_{i, k}\neq0\}|}$ is the probability to choose this pair in our sampling strategy.}
\end{enumerate}
This way, a single training iteration with $b\cdot n$ batch size requires only $O(b)$ runs of Dijkstra algorithm.

After computing the gradients for a minibatch, we update GraphGlove parameters using Adam~\cite{adam} with learning rate $\alpha{=}0.01$ and standard hyperparameters ($\beta_1{=}0.9, \beta_2{=}0.999$).

It took us less than 3.5 hours on a 32-core CPU to train GraphGlove on 50k tokens until convergence. This is approximately 3 times longer than Euclidean GloVe in the same setting.

\section{Experiments}
\label{sect:experiments}

In the main text, we report results for 50k vocabulary with 20 parameters per token. Results for other settings, as well as the standard deviations, can be found in the supplementary material.

\subsection{Word Similarity}
\label{sect:word_similarity}

To measure similarity of a pair of words, we use cosine distance for Euclidean GloVe, the hyperbolic distance for Poincar\'e GloVe and the shortest path distance for GraphGlove. In the main experiments, we exclude pairs with out-of-vocabulary (OOV) words. In the supplementary material, we also provide results with inferred distances for OOV words.

We evaluate word similarity on standard benchmarks: WS353, SCWS, RareWord, SimLex and SimVerb. These benchmarks evaluate Spearman rank correlation of human-annotated similarities between pairs of words and model predictions\footnote{We use standard evaluation code from \url{https://github.com/kudkudak/word-embeddings-benchmarks}}. Table~\ref{tab:sim50k_table} shows that GraphGlove outperforms vector-based embeddings by a large margin.

\begin{table}[t!]
\centering
\begin{tabular}{llccccc}
\toprule
 & & \!\!\!\!\!\!\!\bf SCWS & \bf WS353  & \bf RW & \bf SL & \bf SV\\
\midrule
\multicolumn{7}{l}{\!\!\!\bf Euclidean }\\
&\!\!\!\!\!\!\! & 54.0 & 46.1 & 31.4 & 20.1 & \phantom{0}8.7\\
\midrule
\multicolumn{7}{l}{\!\!\!\bf Poincar\'e }\\
&\!\!\!\!\!\!\! $d^2$ & 45.5 & 41.0 & 33.7 & 23.0 & 10.9 \\
&\!\!\!\!\!\!\! $\cosh^2 d$ & \underline{53.5} & 51.3 & \underline{36.1} & 23.5 & \underline{11.6} \\
\midrule
\multicolumn{7}{l}{\!\!\!\bf Graph}\\
&\!\!\!\!\!\!\! $d$ & \textbf{56.2} & \underline{56.7} & \textbf{37.2} & \textbf{30.4} & 10.3 \\
&\!\!\!\!\!\!\! $\langle \cdot, \cdot\rangle$ & 53.4 & \textbf{58.6} & 35.5 & \underline{30.0} & \textbf{14.4} \\
\bottomrule
\end{tabular}
\caption{Spearman rank correlation on word similarity tasks; best is bold, second best is underlined. 50k vocabulary, 20 parameters per token. Results for other setups can be found in the supplementary material.}
\label{tab:sim50k_table} 
\end{table}

\subsection{Word Analogy}

Analogy prediction is a standard method for evaluation of word embeddings. This task typically contains tuples of 4 words: $(a, a^*, b, b^*)$ such that $a$ is to $a^*$ as $b$ is to $b^*$. The model is tasked to predict $b^*$ given the other three words: for example, ``$a=Athens$ is to $a^*=Greece$ as $b=Berlin$ is to $b^*= \underline{\hspace{1cm}}  (Germany)$''. Models are compared based on accuracy of their predictions across all tuples in the benchmark.

\paragraph{Datasets.} We use two test sets: standard benchmarks~\cite{word2vec,mikolov-etal-2013-linguistic} and the Bigger Analogy Test Set (BATS)~\cite{GladkovaDrozd2016}.

\textit{The standard benchmarks} contain Google analogy~\cite{mikolov2013efficient} and MSR~\cite{mikolov-etal-2013-linguistic} test sets. MSR test set contains only morphological category; Google test set
contains 9 morphological and 5 semantic categories, with 20~--70 unique word pairs per category combined in all possible ways to yield 8,869 semantic and 10,675 syntactic questions. 
Unfortunately, these test sets are not balanced in terms of linguistic relations, 
which may lead to overestimation of analogical reasoning abilities as a whole~\cite{GladkovaDrozd2016}.\footnote{For example, 56.7$\%$ of semantic questions in the Google dataset exploit the same \textit{capital:country} relation, and the MSR dataset only concerns morphological relations.}  %

\textit{The Bigger Analogy Test Set (BATS)}~\cite{GladkovaDrozd2016}  contains 40 linguistic relations, each represented with 50 unique word pairs, making up 99,200 questions in total. In contrast to the standard benchmarks, BATS is balanced across four groups: \textit{inflectional} and \textit{derivational} morphology, and \textit{lexicographic} and \textit{encyclopedic} semantics.

\paragraph{Evaluation.} Euclidean GloVe solves analogies by maximizing the 3\textsc{CosAdd} score:
\vspace{-1ex}
$$
    b^*{=}\!\!\underset{\hat b \in V \setminus \{a^*, a, b\}}{\arg\!\max}\!\!\left( {\cos(\hat b, a^*){-}\cos(\hat b, a){+}\cos(\hat b, b)}\right).
$$
We adapt this for GraphGlove by substituting $\cos(x, y)$ with a graph-based similarity function. As a simple heuristic, we define the similarity between two words as the correlation of vectors consisting of distances to all words in the vocabulary:
\vspace{-1ex}
    $$\vec{d}_G(x) = (d_G(x, v_0), ..., d_G(x, v_N))$$
    \vspace{-2ex}
    $$\text{sim}_G(x, y) := \text{corr}(\vec{d}_G(x), \vec{d}_G(y))$$
This function behaves similarly to the cosine similarity: its values are from -1 to 1, with unrelated words having similarity close to 0 and semantically close words having similarity close to 1. Another alluring property of $\text{sim}_G(x, y)$ is efficient computation: we can get full distance vector $\vec{d}_G(x)$ with a single pass of Dijkstra's algorithm.%

We use $\text{sim}_G(x, y)$ to solve the analogy task in GraphGlove:
\vspace{-1ex}
$$
    b^*{=}\!\!\underset{\hat b \in V \setminus \{a^*, a, b\}}{\!\arg\!\max}\!\!\!\left(\! {\text{sim}(\hat b, a^*)} {-} \text{sim}(\hat b, a) {+} \text{sim}(\hat b, b)\right).
$$

For details on how Poincar\'e GloVe solves the analogy problem, we refer the reader to the original paper~\cite{tifrea2018poincar}.

\paragraph{Results.} GraphGlove shows substantial improvements over vector-based baselines (Tables~\ref{tab:analogy_table} and~\ref{tab:bats}). Note that for Poincar\'e GloVe, the best-performing loss functions for the two tasks are different ($\cosh^2d$ for similarity and $d^2$ for analogy), and there is no setting where Poincar\'e GloVe outperforms Euclidean Glove on both tasks. While for GraphGlove best-performing loss functions also vary across tasks, GraphGlove with the dot product loss outperforms all vector-based embeddings on 10 out of 13 benchmarks (both analogy and similarity). This shows that when removing limitations imposed by the geometry of a vector space, embeddings can better reflect the structure of the data.  We further confirm this by analyzing the properties of the learned graphs in Section~\ref{sect:analysis}.

\begin{table}[t!]
\centering
\begin{tabular}{llcccc}
\toprule
 & & \bf Sem.& \bf Syn.  & \bf Full & \bf MSR\\

\midrule
\multicolumn{6}{l}{\bf Euclidean }\\
& & 30.8 & 20.9 & 25.2 & 15.5 \\
\midrule
\multicolumn{6}{l}{\bf Poincar\'e }\\
& $d^2$ & \underline{31.5} & 20.3 & \underline{25.4} & \underline{19.7} \\
& $\cosh^2 d$ & 30.5 & 16.9 & 23.1 & 18.1 \\ 
\midrule
\multicolumn{6}{l}{\bf Graph}\\
& $d$ & 31.3 & \underline{20.5} & \underline{25.4} & 16.1 \\
& $\langle \cdot, \cdot\rangle$ & \textbf{33.0} & \textbf{24.2} & \textbf{28.2} & \textbf{21.7} \\
\bottomrule
\end{tabular}
\caption{Accuracy word analogy tasks; best is bold, second best is underlined. 50k vocabulary, 20 parameters per token. Results for other setups can be found in the supplementary material. (\textsc{Sem., Syn.} and \textsc{Full} are Google benchmarks~\cite{mikolov2013efficient}).}
\label{tab:analogy_table}
\end{table}

\begin{table}[t!]
\centering
\begin{tabular}{llcccc}
\toprule
 & & \!\!\!\!\!\!\!\bf Inf. & \bf Der.  & \bf Lex.&\bf Enc.\\
\midrule
\multicolumn{6}{l}{\!\!\!\bf Euclidean }\\
&\!\!\!\!\!\!\! & 14.3&2.1&18.3&3.7 \\
\midrule
\multicolumn{6}{l}{\!\!\!\bf Poincar\'e }\\
&\!\!\!\!\!\!\! $d^2$ & 14.8&\underline{2.3}&18.9&4.3 \\
&\!\!\!\!\!\!\! $\cosh^2 d$& 15.7&\textbf{2.4}&19.2&4.4 \\
\midrule
\multicolumn{6}{l}{\!\!\!\bf Graph}\\
&\!\!\!\!\!\!\! $d$ & \underline{15.9}&2.3&\underline{19.3}&\underline{4.6}\\
&\!\!\!\!\!\!\! $\langle \cdot, \cdot\rangle$ & \textbf{16.9}&2.2&\textbf{20.6}&\textbf{5.4} \\
\bottomrule
\end{tabular}
\caption{Spearman rank correlation on BATS word analogy dataset; best is bold, second best is underlined. 50k vocabulary, 20 parameters per token.}
\label{tab:bats} 
\end{table}

\section{Learned Graph Structure}
\label{sect:analysis}

In this section, we analyze the graph structure learned by our method and reveal its differences from the structure of vector-based embeddings. 

We compare graph $G_{G}$ learned by GraphGlove~($d$)  with graphs $G_E$ and $G_P$ induced from Euclidean and Poincar\'e~($\cosh^2d$) embeddings respectively.\footnote{We take the same models as in Section~\ref{sect:experiments}.}
For vector embeddings, we consider two methods of graph construction: 
\begin{enumerate}[noitemsep]
    \item \textsc{thr}~-- connect two nodes if they are closer than some threshold $\tau$,
    \item \textsc{knn}~-- connect each node to its $K$ nearest neighbors and combine multiple edges.
\end{enumerate}
The values $\tau$ and $K$ are chosen to have similar edge density for all graphs.\footnote{Namely, $K=13$ and  $\tau=0.112$ for Euclidean GloVe, $K=13$ and $\tau=0.444$ for Poincar\'e Glove.}

We find that in contrast to the graphs induced from vector embeddings:
\begin{itemize}
    \item in GraphGlove frequent and generic words are highly interconnected;
    \item GraphGlove has hierarchical, similar to WordNet, structure;
    \item GraphGlove has non-trivial geometry containing subgraphs with different local topology.
\end{itemize}

\subsection{Important words}

Here we identify which words correspond to ``central'' (or important) nodes in different graphs; we consider several notions of node centrality frequently used in graph theory. Note that in this section, by word importance we mean graph-based properties of nodes (e.g. the number of neighbors), and not semantic importance (e.g., high importance for content words and low for function words).

\paragraph{Degree centrality.}

\begin{figure}[t!]
    \centering
    \includegraphics[width=\columnwidth]{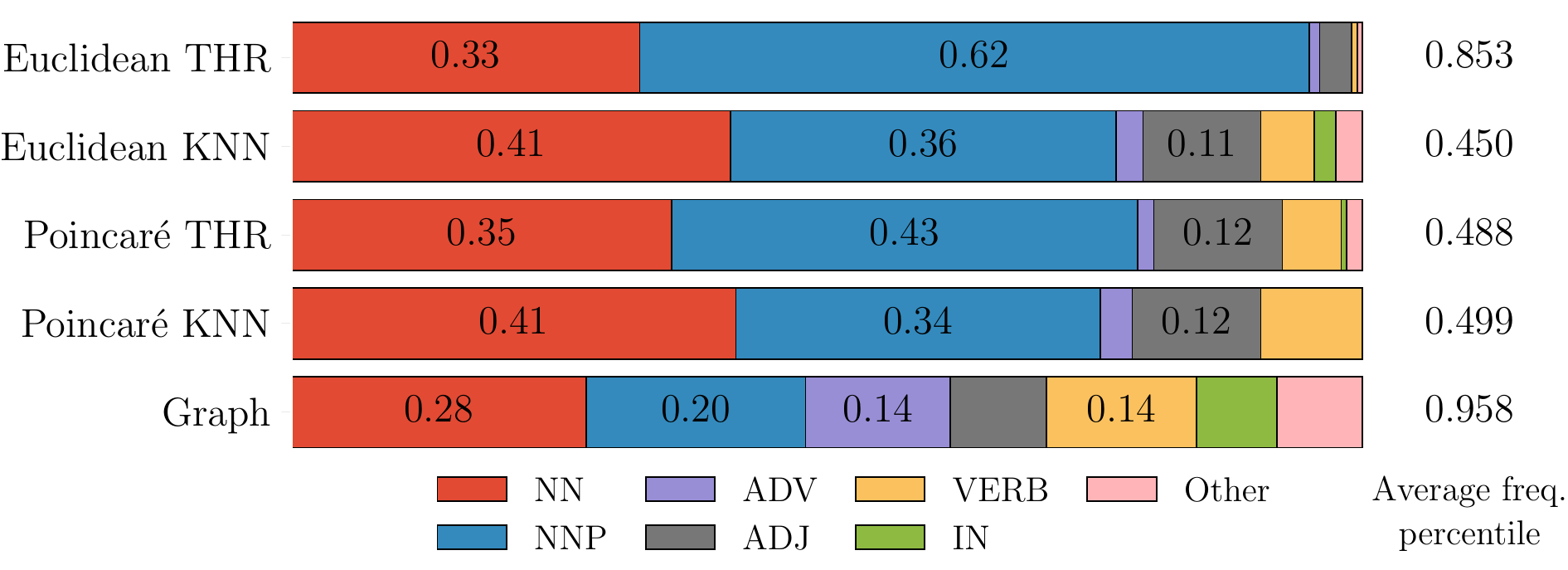}
    \caption{Top-200 words, the degree centrality. POS distribution and the average frequency percentile.}
    \label{fig:degree}
    \vspace{-1ex}
\end{figure}{}

The simplest measure of node importance is its degree. 
For the top 200 nodes with the highest degree, we show the distribution of parts of speech and the average frequency percentile (higher means more frequent words). Figure~\ref{fig:degree} shows that for all vector-based graphs, the top contains a significant fraction of proper nouns 
and nouns. For $G_G$, distribution of parts of speech is more uniform and the words are more frequent. We provide the top words %
and all subsequent importance measures in the supplementary material.

\paragraph{Eigenvector centrality.}

\begin{figure}[t!]
    \centering
    \includegraphics[width=\columnwidth]{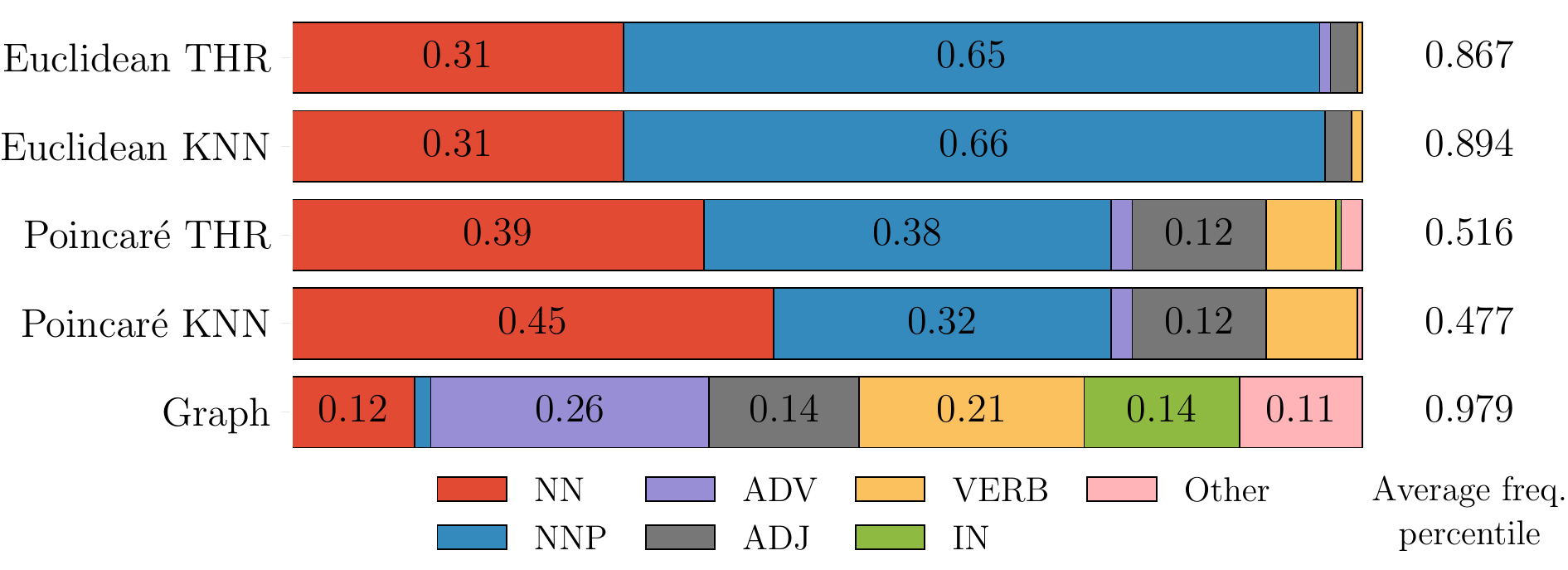}
    \caption{Top-200 words, the eigenvector centrality. POS distribution and the average frequency percentile.}
    \label{fig:eigen}
    \vspace{-2ex}
\end{figure}{}

A more robust measure of node importance is the eigenvector centrality~\citep{bonacich1987power}. This centrality takes into account not only the degree of a node but also the importance of its neighbors: a high eigenvector score means that a node is connected to many nodes who themselves have high scores.

Figure~\ref{fig:eigen} shows that for $G_{G}$ the top changes in a principled way: the average frequency increases, proper nouns almost vanish, many adverbs, prepositions, linking and introductory words appear (e.g., `well', `but', `in', `that').\footnote{See the words in the supplementary material.} For $G_G$, the top consists of frequent generic words; this agrees with the intuitive understanding of importance. %
Differently from $G_G$, top words for $G_E$ and $G_P$ have lower frequencies, fewer adverbs and prepositions. This can be because it is hard to make generic words from different areas close for vector-based embeddings, while GraphGlove can learn arbitrary connections. 

\paragraph{$k$-core.}

\begin{table}[t!]
\centering
\begin{tabular}{lcc}
\toprule
 & size & $k$ \\
\midrule
\textbf{Euclidean} & 275 & 198 \\
\textbf{Poincar\'e} & 235 & 156 \\
\textbf{Graph} & 197 & 21 \\
\bottomrule
\end{tabular}
\vspace{-1ex}
\caption{The main core size and its $k$ value. For vector-based embeddings, the $\textsc{thr}$ graphs are shown (by construction, the main core of a $\textsc{knn}$ graph is trivial).}
\label{tab:core_size_k}
\vspace{-2ex}
\end{table}

To further support this claim, we looked at the main $k$-core of the graphs. Formally, \textit{$k$-core} is a maximal subgraph that contains nodes of degree $k$ or more; \textit{the main core} is non-empty core with the largest $k$. Table~\ref{tab:core_size_k} shows the sizes of the main cores and the corresponding values of $k$.
 Note that the maximum $k$ is much smaller for $G_G$; a possible explanation is that the cores in $G_{E}$ and $G_P$ are formed by nodes in highly dense regions of space, while in $G_G$ the most important nodes in different parts can be interlinked together.

\subsection{The Structure is Hierarchical}

In this section, we show that the structure of our graph reflects the hierarchical nature of words. We do so by comparing the structure learned by GraphGlove to the noun hierarchy from WordNet.
To extract hierarchy from $G_G$, we (1)~take all (lemmatized) nouns in our dataset which are also present in WordNet (22.5K words), (2)~take the root noun `entity' (which is the root of the WordNet tree), and (3)~construct the hierarchy: the $k$-th level is formed by all nodes at edge distance %
$k$ from the root.

We consider two ways of measuring the agreement between the hierarchies: \textit{word correlation} and \textit{level correlation}. Word correlation is Spearman's rank correlation between the vectors of levels for all nouns. Level correlation is Spearman's rank correlation between the vectors $l$ and $l^{avg}$, where $l_i$ is the level in WordNet tree and $l_i^{avg}$ is the average level of $l_i$'s words in our hierarchy.

We performed these measurements for all graphs (see Table~\ref{tab:hierarchy}).\footnote{The low performance of threshold-based graphs can be explained by the fact that they are highly disconnected (we assume that all nodes which are not connected to the root form the last level).} We see that, according to both correlations, $G_G$ is in better agreement with the WordNet hierarchy.

\begin{table}[t!]
\centering
\begin{tabular}{lccc}
\toprule
& & Word & Level \\
& & correlation & correlation \\
\midrule
{\bf Euclidean}&\textsc{thr} &0.016&0.118\\
&\textsc{knn} &0.149&0.539\\
\midrule
{\bf Poincar\'e}&\textsc{thr} &0.018&0.122\\
&\textsc{knn} &0.124&0.094\\
\midrule
{\bf Graph}&& \textbf{0.199} & \textbf{0.650} \\
\bottomrule
\end{tabular}
\vspace{-1ex}
\caption{Correlations of hierarchies extracted from graphs and WordNet levels.}
\label{tab:hierarchy}
\vspace{-2ex}
\end{table}

\begin{figure*}[t!]
    \centering
     \subfloat[$\delta \approx 0$]{\includegraphics[width=.6\columnwidth]{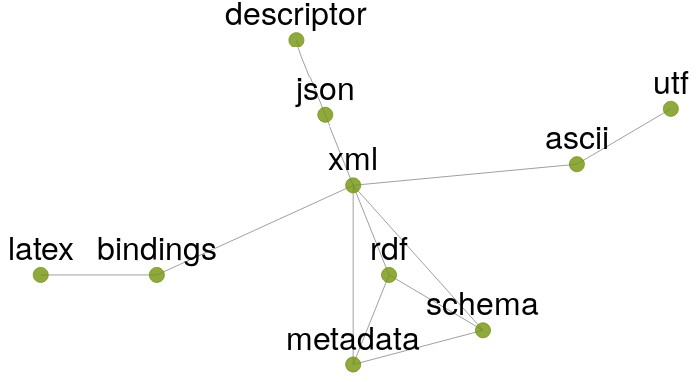}}\quad\quad
    \subfloat[$\delta \approx 0.15$]{\includegraphics[width=.6\columnwidth]{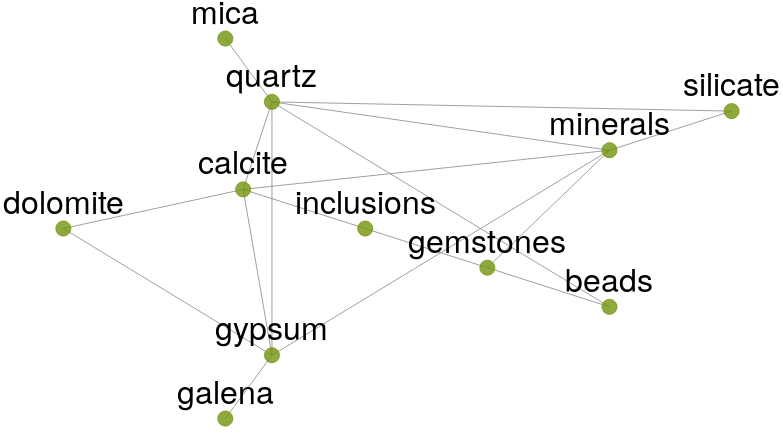}}\quad\quad
    \subfloat[$\delta \approx 0.35$]{\includegraphics[width=.6\columnwidth]{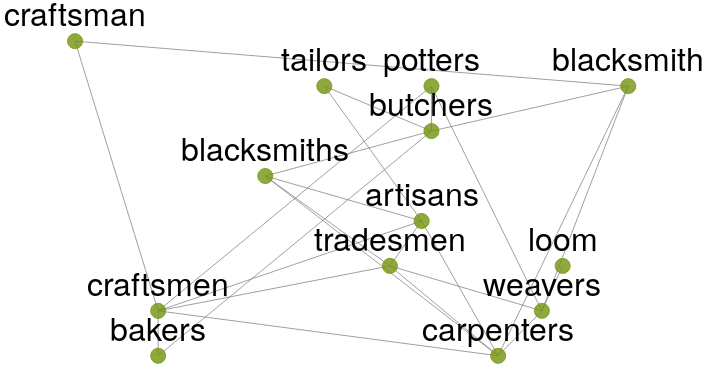}}
    \vspace{-1ex}
    \caption{Examples of clusters with various $\delta$-hyperbolicities. For more interactive cluster visualizations, visit \url{https://yandex-research.github.io/graph-glove/} }
    \label{fig:clusters}
    \vspace{-2ex}
\end{figure*}

\subsection{The Geometry is Non-trivial}

In contrast to vector embeddings, graph-based representations are not constrained by a vector space geometry and potentially can imitate arbitrarily complex spaces. Here we confirm that the geometry learned by GraphGlove is indeed non-trivial.

We cluster $G_G$ using the Chinese Whispers algorithm for graph node clustering~\citep{biemann2006chinese} and measure Gromov $\delta$-hyperbolicity for each cluster. Gromov hyperbolicity measures how close is a given metric to a tree metric (see, e.g., \citet{tifrea2018poincar} for the formal definition) and has previously been used to show the tree-like structure of the word log-co-occurrence graph~\cite{tifrea2018poincar}. Low average $\delta$ indicates tree-like structure with $\delta$ being exactly zero for trees; $\delta$ is usually normalized by the average shortest path length to get a value invariant to metric scaling.

\begin{figure}[t!]
\vspace{-1ex}
    \centering
    \includegraphics[width=0.85\columnwidth]{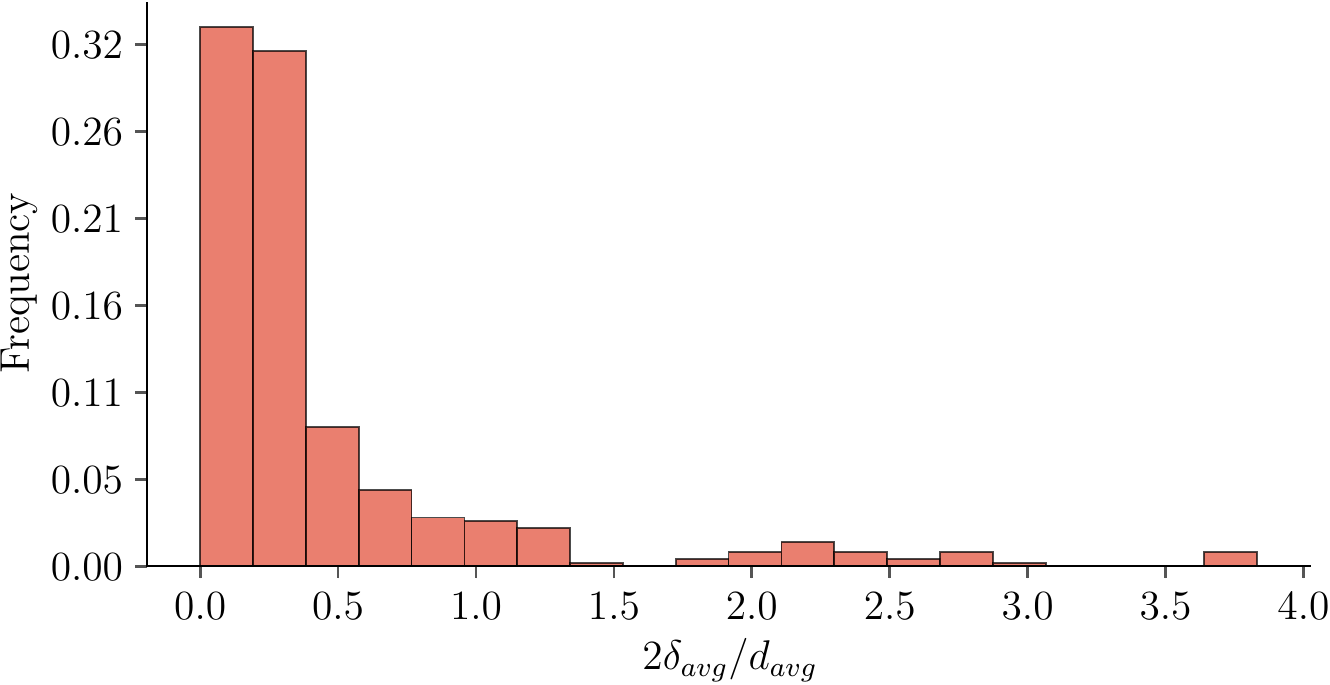}
    \vspace{-1ex}
    \caption{Distribution of cluster $\delta$-hyperbolicities, normalized by the average shortest path length in a cluster.}
    \vspace{-2ex}
    \label{fig:gromov_hyp}
\end{figure}

Figure~\ref{fig:gromov_hyp} shows the distribution of average $\delta$-hyperbolicity for clusters of size at least 10. Firstly, we see that for many clusters the normalized average $\delta$-hyperbolicity is close to zero, which agrees with the intuition that some words form a hierarchy.
Secondly, $\delta$-hyperbolicity varies significantly over the clusters and some clusters have relatively large values; it means that these clusters are not tree-like. Figure~\ref{fig:clusters} shows examples of clusters with different values of $\delta$-hyperbolicity: both tree-like~(Figure~\ref{fig:clusters}a) and more complicated~(Figure~\ref{fig:clusters}b-c).

\section{Related Work}

Word embedding methods typically represent words as vectors in a low-dimensional space; usually, the vector space is Euclidean~\cite{word2vec,pennington-etal-2014-glove,bojanowski-etal-2017-enriching}, but recently %
other spaces, e.g. hyperbolic, have been explored~\cite{leimeister2018skip,dhingra2018embedding,tifrea2018poincar}. 
However, vectorial embeddings %
can have undesired properties:
 e.g., in dot product spaces certain words cannot be assigned high probability regardless of their context~\cite{demeter2020stolen}.
A conceptually different approach is to model words as probability density functions~\cite{vilnis2014word,athiwaratkun-wilson-2017-multimodal,brazinskas-etal-2018-embedding,muzellec2018generalizing,athiwaratkun2018on}. We propose a new setting: embedding words as nodes in a weighted graph. 

Representing language data in the form of a graph has been a long-standing task~\cite{miller1995wordnet,motter2002topology,cancho2001small,niyogi2006computational,masucci2006network}. Graph lexicons were used to learn word embeddings specialized towards certain types of lexical knowledge~\cite{nguyen-etal-2017-hierarchical,vulic-mrksic-2018-specialising,liu-etal-2015-learning,ono-etal-2015-word,mrksic-etal-2017-semantic,bollegala2016joint}. It is also possible to incorporate external linguistic information from graphs, e.g. dependency parser outputs~\cite{vashishth2018incorporating}.

To learn a weighted graph, we use the method by~\citet{PRODIGE}. 
Prior approaches to learning graphs from data are eigher highly problem-specific and not scalable \citet{escolano2011points,karasuyama2017adaptive,kang2019robust} or solve a less general but important case of learning \textit{directed acyclic graphs}~\cite{zheng2018dags,yu2019dag}. 
The opposite to learning a graph from data is the task of embedding nodes in a given graph to reflect graph distances and/or other properties; see \citet{hamilton2017representation} for a thorough survey.

Analysis of word embeddings and the structure of the learned feature space often reveals interesting language properties and is an important research direction~\cite{kohn-2015-whats,bolukbasi2016man,mimno-thompson-2017-strange,nakashole-flauger-2018-characterizing,naik-etal-2019-exploring,ethayarajh-etal-2019-understanding}. We show that graph-based embeddings can be a powerful tool for language analysis.

\section{Conclusions}

We introduce GraphGlove~--- graph word embeddings, where each word is a node in a weighted graph and the distance between words is the shortest path distance between the corresponding nodes. The graph is learned end-to-end in an unsupervised manner. We show that GraphGlove substantially outperforms both Euclidean and Poincar\'e GloVe on word similarity and word analogy tasks. Our analysis reveals that the structure of the learned graphs is hierarchical and similar to that of WordNet; the geometry is highly non-trivial and contains subgraphs with different local topology. 

Possible directions for future work include using GraphGlove for unsupervised hypernymy detection, analyzing undesirable word associations, comparing learned graph topologies for different languages, and downstream applications such as sequence classification. Also, given the recent success of models such as ELMo and BERT, it would be interesting to explore extensions of GraphGlove to the class of contextualized embeddings.

\section*{Acknowledgments}
We would like to thank Artem Babenko for inspiring the authors to work on this paper. We also thank the anonymous reviewers for valuable feedback. %

\bibliographystyle{acl_natbib}
\bibliography{emnlp2020}

\newpage

\appendix

\section{Appendix: Additional benchmarks}

\subsection{Variance study}
\begin{figure*}[t!]
    \centering
    \includegraphics[width=\linewidth]{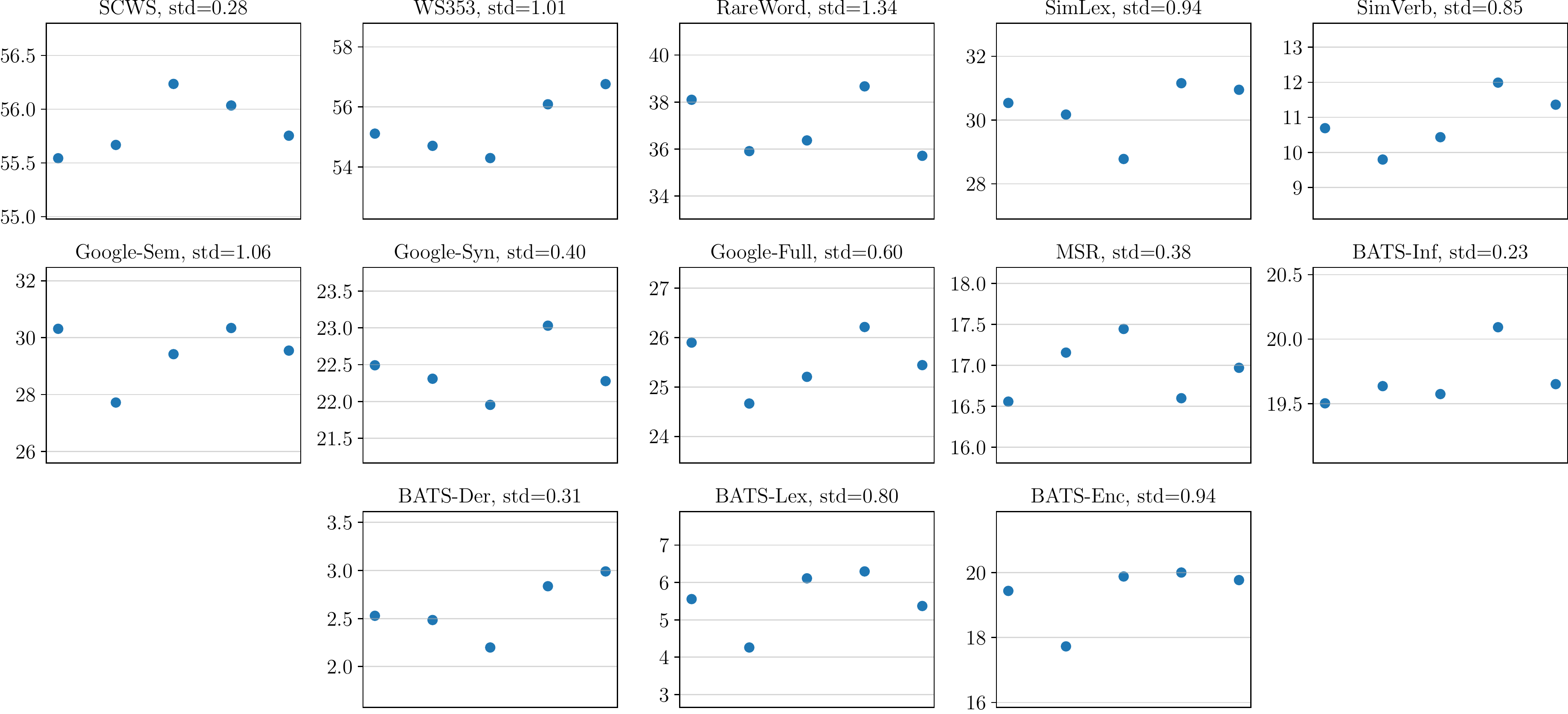}
    \caption{Results of evaluation for GraphGlove with 20 parameters per vertex on all benchmarks over 5 random initializations. Here, $d$ is used in the loss function.}
    \label{fig:variance}
\end{figure*}

As our method relies on random initialization of a graph in PRODIGE, a natural question is whether different choice of drawn edges significantly affects the quality of representations in the end of training. Figure \ref{fig:variance} demonstrates that after running the training procedure with distance-based loss for 5 different random seeds, the final metrics values have a standard deviation of less than 1 point in 10/13 tasks and have a standard deviation of at most 1.34 percent for the RareWord dataset. Thus, we can conclude that GraphGlove results are relatively stable with respect to selection of random edges  before training.

{}

\subsection{Similarity}
Below we report additional similarity benchmarks for GraphGlove and its vectorial counterparts:
\begin{itemize}
    \item 50K tokens, 100 parameters / token - Table \ref{tab:sim50k_100d_table};
    \item 200K tokens, 20 parameters / token - Table \ref{tab:sim200k_20d_table}.
\end{itemize}

Some word pairs in each similarity benchmark are out of vocabulary (OOV). In the main evaluation, we drop such pairs from the benchmark. However, there's also a different way to deal with such words. 

A popular workaround is to calculate the distance between $w_i$ and OOV as an average distance from $w_i$ to other words. In the rare case when both words are OOV, we can consider them infinitely distant from each other. We report similarity benchmarks including OOV tokens in Tables \ref{tab:sim50k_20d_table_infer}, \ref{tab:sim50k_100d_table_infer} and \ref{tab:sim200k_20d_table_infer}.

\begin{table}[t!]
\centering%

\begin{tabular}{llccccc}
\toprule
 & & \!\!\!\!\!\!\!\bf SCWS & \bf WS353  & \bf RW & \bf SL & \bf SV\\

\midrule
\multicolumn{7}{l}{\!\!\!\bf Euclidean }\\
&\!\!\!\!\!\!\! & 58.0 & 62.0 & 38.3 & 28.9 & 12.4 \\
\midrule
\multicolumn{7}{l}{\bf Poincar\'e }\\
&\!\!\!\!\!\!\! $d^2$ 
  & 53.2 & 57.3 & 40.8 & 29.4 & 12.56 \\
&\!\!\!\!\!\!\! $\cosh^2 d$
  & \bf 59.5 & \bf 65.9 & 45.8 & 31.6 & 13.7 \\
\midrule
\multicolumn{7}{l}{\bf Graph}\\
&\!\!\!\!\!\!\! $d$ 
  & 59.3 & 65.5 & \bf 46.0 & \bf 33.6 & 12.6 \\
&\!\!\!\!\!\!\! $\langle \cdot, \cdot\rangle$ 
  & 56.8 & 61.6 & 42.6 & 32.8 & \bf 14.8 \\
\bottomrule
\end{tabular}
\caption{Spearman rank correlation on word similarity tasks. Models  with 50k vocab., 100 parameters per token. Skip word pairs that contain OOV words.}
\label{tab:sim50k_100d_table}
\end{table}

\begin{table}[t!]
\centering%

\begin{tabular}{llccccc}
\toprule
 & & \!\!\!\!\!\!\!\bf SCWS & \bf WS353  & \bf RW & \bf SL & \bf SV\\

\midrule
\multicolumn{7}{l}{\!\!\!\bf Euclidean }\\
&\!\!\!\!\!\!\! & 55.6 & 49.9 & 31.8 & 23.6 & 10.6 \\
\midrule
\multicolumn{7}{l}{\bf Poincar\'e }\\
&\!\!\!\!\!\!\! $d^2$ 
    & 46.2 & 45.0 & 28.5 & 23.5 & 12.0 \\
&\!\!\!\!\!\!\! $\cosh^2 d$
    & 56.1 & 51.2 & 32.0 & \bf 24.8 & 12.4 \\
\midrule
\multicolumn{7}{l}{\bf Graph}\\
&\!\!\!\!\!\!\! $d$ 
    & \bf 58.5 & \bf 56.4 & \bf 33.4 & 23.4 & 11.5\\
&\!\!\!\!\!\!\! $\langle \cdot, \cdot\rangle$ 
    & 53.2 & 52.9 & 30.9 & 23.4 & \bf 13.8 \\
\bottomrule
\end{tabular}
\caption{Spearman rank correlation on word similarity tasks. Models with 200k vocab., 20 parameters per token. Ignore word pairs with OOV words.}
\label{tab:sim200k_20d_table}
\end{table}

\begin{table}[t]
\centering
\begin{tabular}{llccccc}
\toprule
 & & \!\!\!\!\!\!\!\bf SCWS & \bf WS353  & \bf RW & \bf SL & \bf SV\\

\midrule
\multicolumn{7}{l}{\!\!\!\bf Euclidean }\\
&\!\!\!\!\!\!\! & 50.5 & 44.9 & 11.5 & 19.5 & 6.7 \\
\midrule
\multicolumn{7}{l}{\bf Poincar\'e }\\
&\!\!\!\!\!\!\! $d^2$
    & 42.8 & 38.8 & 8.4 & 21.8 & 7.8 \\
&\!\!\!\!\!\!\! $\cosh^2 d$
    & 49.8 & 49.8 & 11.6 & 22.3 & 8.3 \\
\midrule
\multicolumn{7}{l}{\bf Graph}\\
&\!\!\!\!\!\!\! $d$ 
    & \bf 51.7 & 55.5 & 7.3 & \bf 30.0 & 8.9 \\
&\!\!\!\!\!\!\! $\langle \cdot, \cdot\rangle$ 
    & 48.6 & \bf 57.9 & \bf 11.8 & 28.8 & \bf 10.7 \\
\bottomrule
\end{tabular}
\caption{Spearman rank correlation on word similarity tasks. Models  with 50k vocab., 20 parameters per token. Infer distances to OOV words. }
\label{tab:sim50k_20d_table_infer}
\end{table}

\begin{table}[h]
\centering%

\begin{tabular}{llccccc}
\toprule
 & & \!\!\!\!\!\!\!\bf SCWS & \bf WS353  & \bf RW & \bf SL & \bf SV\\

\midrule
\multicolumn{7}{l}{\!\!\!\bf Euclidean }\\
&\!\!\!\!\!\!\! & 54.2	& 58.8 & 9.2 & 27.8 & 8.8 \\
\midrule
\multicolumn{7}{l}{\bf Poincar\'e }\\
&\!\!\!\!\!\!\! $d^2$ 
    & 50.3 & 54.5 & 10.3 & 27.7 & 9.9 \\
&\!\!\!\!\!\!\! $\cosh^2 d$
    & 55.4 & \bf 60.1 & \bf 12.8 & 29.6 & 10.0 \\
\midrule
\multicolumn{7}{l}{\bf Graph}\\
&\!\!\!\!\!\!\! $d$ 
    & \bf 55.7 & 60.0 & 12.8 & \bf 32.8 & 9.9 \\
&\!\!\!\!\!\!\! $\langle \cdot, \cdot\rangle$ 
    & 54.0 & 58.5 & 10.7 & 31.0 & \bf 11.2 \\
\bottomrule
\end{tabular}
\caption{Spearman rank correlation on word similarity tasks. Models  with 50k vocab., 100 parameters per token. Infer distances to OOV words.}%
\label{tab:sim50k_100d_table_infer}
\end{table}

\begin{table}[t!]
\centering%

\begin{tabular}{llccccc}
\toprule
 & & \!\!\!\!\!\!\!\bf SCWS & \bf WS353  & \bf RW & \bf SL & \bf SV\\

\midrule
\multicolumn{7}{l}{\!\!\!\bf Euclidean }\\
&\!\!\!\!\!\!\! & 54.4 & 49.9 & 24.9 & 23.7 & 10.8 \\
\midrule
\multicolumn{7}{l}{\bf Poincar\'e }\\
&\!\!\!\!\!\!\! $d^2$ 
    & 46.0 & 45.0 & 23.8 & 23.9 & 11.8 \\
&\!\!\!\!\!\!\! $\cosh^2 d$
    & 56.0 & 51.2 & 27.5 & \bf 24.9 & 12.3 \\
\midrule
\multicolumn{7}{l}{\bf Graph}\\
&\!\!\!\!\!\!\! $d$ 
    & \bf 58.1 & \bf 56.4 & \bf 28.6 & 23.6 & 11.3 \\
&\!\!\!\!\!\!\! $\langle \cdot, \cdot\rangle$ 
    & 53.0 & 52.9 & 26.2 & 23.9 & \bf 13.5 \\
\bottomrule
\end{tabular}
\caption{Spearman rank correlation on word similarity tasks. Models with 200k vocab., 20 parameters per token. Infer distances to missing words.}%
\label{tab:sim200k_20d_table_infer}
\end{table}

\subsection{Analogy}
We also evaluate these scenarios for Analogy Prediction:
\begin{itemize}
    \item 50K tokens, 100 parameters / token - Table \ref{tab:analogy50k_100d_table};
    \item 200K tokens, 20 parameters / token - Table \ref{tab:analogy200k_20d_table};
\end{itemize}

\begin{table}[t]\renewcommand{\arraystretch}{1.25}
\centering

\begin{tabular}{lcccc}
\toprule
  & \textbf{Sem.} & \textbf{Syn.} & \textbf{Full} & \textbf{MSR} \\
\midrule
\bf Euclidean \\
& 61.6 & 50.0 & 55.1 & 48.0 \\
\midrule
\multicolumn{5}{l}{\bf Poincar\'e }\\
$d^2$ & 49.4 & 38.3 & 43.2 & 26.1 \\ 
$\cosh^2 d$ & \bf 63.2 & 56.8 & 59.6 & 47.8 \\
\midrule
\multicolumn{5}{l}{\bf Graph }\\
$d$ & 59.9 & 58.3 & 59.0 & 46.0 \\
$\langle\cdot,\cdot\rangle$ & 61.8 & \bf 59.8 & \bf 60.7 & \bf 49.7 \\
\bottomrule
\end{tabular}
\caption{Analogy prediction accuracy. Models with 50K tokens, 100 parameters per token.}%
\label{tab:analogy50k_100d_table}
\end{table}

\begin{table}[t]\renewcommand{\arraystretch}{1.25}
\centering

\begin{tabular}{lcccc}
\toprule
  & \textbf{Sem.} & \textbf{Syn.} & \textbf{Full} & \textbf{MSR} \\
\midrule
\bf Euclidean \\
& 31.7 & 19.7 & 25.2 & 11.6 \\

\midrule
\multicolumn{5}{l}{\bf Poincar\'e }\\
$d^2$ & 32.9 & 20.9 & 26.4 & 14.4 \\ 
$\cosh^2 d$ & 31.2 & 19.9 & 25.0 & 14.0 \\
\midrule
\multicolumn{5}{l}{\bf Graph }\\
$d$ & 32.6 & 19.7 & 25.6 & 13.5 \\
$\langle\cdot,\cdot\rangle$ & \bf 34.3 & \bf 24.7 & \bf 29.1 & \bf 19.7 \\
\bottomrule
\end{tabular}
\caption{Analogy prediction accuracy. Models with 200K tokens, 20 parameters per token.}%
\label{tab:analogy200k_20d_table}
\end{table}

\section{Supplementary material: graph central nodes}

\paragraph{Top 20 words by degree centrality}

\begin{itemize}
\item Euclidean THR:
['cummings', 'glover', 'boyd', 'hooper', 'barrett', 'hicks', 'mckay', 'dunn', 'kemp', 'moran', 'payne', 'ingram', 'harrington', 'webb', 'ellis', 'jenkins', 'goodwin', 'benson', 'corbett', 'willis']
\item Euclidean KNN:
['bunn', 'willey', 'cottrell', 'sandys', 'alfaro', 'forgets', 'ellis', 'minaj', 'taylor', 'lemaire', 'lockwood', 'amused', 'emiliano', 'mckay', 'boyd', 'hurtado', 'wonderfully', 'russell', 'this', 'mundy']
\item Poincar\'e THR:
['mundy', 'merriman', 'hoskins', 'cottrell', 'oakes', 'mayne', 'griggs', 'bunn', 'hooper', 'munn', 'gillies', 'glanville', 'beal', 'bartley', 'halloran', 'mcnab', 'purdy', 'bullard', 'willett', 'roper']
\item Poincar\'e KNN:
['imc', 'cottrell', 'willett', 'foxy', 'heim', 'noa', 'mundy', 'newland', 'bunn', 'importantly', 'krug', 'grips', 'hooper', 'haney', 'mcnab', 'misplaced', 'doty', 'taki', 'rushton', 'likewise']
\item Graph:
['bennett', 'even', 'same', 'allen', 'james', 'this', 'although', 'howard', 'however', 'particular', 'example', 'wilson', 'robinson', 'rather', 'well', 'only', 'furthermore', 'fact', 'beginning', 'smith']
\end{itemize}

\paragraph{Top 20 words by eigenvector centrality}

\begin{itemize}
\item Euclidean THR:
['dunn', 'hooper', 'boyd', 'barrett', 'jenkins', 'ellis', 'hicks', 'webb', 'payne', 'cummings', 'benson', 'kemp', 'willis', 'glover', 'mckay', 'moran', 'phillips', 'steele', 'chapman', 'roberts']
\item Euclidean KNN:
['taylor', 'ellis', 'russell', 'benson', 'phillips', 'thompson', 'robinson', 'moore', 'roberts', 'stevens', 'allen', 'curtis', 'webb', 'willis', 'harvey', 'chapman', 'steele', 'jones', 'smith', 'boyd']
\item Poincar\'e THR:
['hoskins', 'oakes', 'hooper', 'gillies', 'roper', 'whitmore', 'corrigan', 'waddell', 'metcalfe', 'goodwin', 'bowles', 'mundy', 'sanderson', 'kemp', 'tobin', 'merriman', 'harrington', 'mccallum', 'cartwright', 'halloran']
\item Poincar\'e KNN:
['willett', 'cottrell', 'bunn', 'rushton', 'doty', 'mundy', 'munn', 'brower', 'rowell', 'glanville', 'macklin', 'purnell', 'mcnab', 'clapp', 'tasker', 'treadwell', 'nichol', 'newland', 'willey', 'prichard']
\item Graph:
['even', 'same', 'however', 'this', 'only', 'although', 'well', 'another', 'both', 'in', 'while', 'rather', 'fact', 'that', 'once', 'though', 'furthermore', 'taken', 'but', 'particular']
\end{itemize}

\paragraph{Main $k$-cores}

\begin{figure}[t!]
    \centering
    \includegraphics[width=\columnwidth]{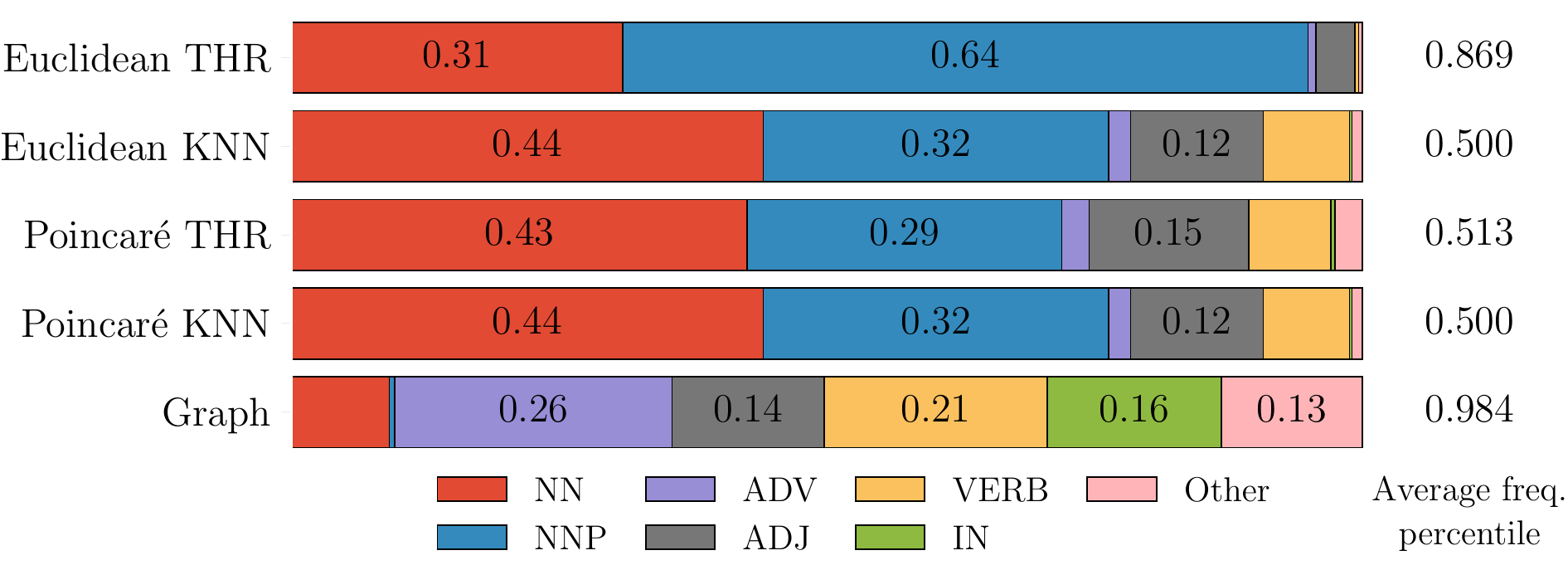}
    \caption{POS distribution for main k-core}
    \label{fig:k_core}
\end{figure}

\begin{itemize}
    \item Euclidean THR:
['nathan', 'harold', 'dick', 'sullivan', 'donaldson', 'yates', 'kerr', 'terry', 'byron', 'duncan', 'horton', 'kelley', 'parker', 'oliver', 'duffy', 'mclean', 'jeremy', 'bartlett', 'osborne', 'howell', 'webb', 'hughes', 'gould', 'pearce', 'bradley', 'walker', 'barry', 'davidson', 'graham', 'jesse', 'blair', 'collins', 'bailey', 'burnett', 'harrington', 'cole', 'wilson', 'jonathan', 'gary', 'tim', 'leslie', 'cunningham', 'elliot', 'jones', 'gorman', 'holt', 'lynch', 'holloway', 'wheeler', 'reid', 'cox', 'evan', 'freeman', 'burgess', 'barker', 'skinner', 'baxter', 'hayes', 'amos', 'colin', 'ritchie', 'harry', 'campbell', 'nolan', 'howard', 'simpson', 'mckenzie', 'ralph', 'gibson', 'rowland', 'lynn', 'jarvis', 'chandler', 'dawson', 'trevor', 'carter', 'stevens', 'nichols', 'hurley', 'hines', 'steele', 'payne', 'shaw', 'phillip', 'henderson', 'richardson', 'andrews', 'anthony', 'briggs', 'roberts', 'fletcher', 'mccall', 'gallagher', 'robertson', 'fowler', 'lowe', 'harris', 'johnson', 'willis', 'carr', 'goodwin', 'spencer', 'neil', 'noel', 'tucker', 'atkins', 'dixon', 'miller', 'bates', 'sweeney', 'barlow', 'andrew', 'wright', 'reeves', 'glover', 'irwin', 'phillips', 'hanson', 'dillon', 'mitchell', 'armstrong', 'farrell', 'stevenson', 'nicholson', 'stewart', 'baker', 'norris', 'coleman', 'hicks', 'foley', 'jack', 'wilkinson', 'powell', 'nelson', 'dalton', 'lewis', 'murray', 'boyd', 'watson', 'elliott', 'hobbs', 'turner', 'horne', 'derek', 'walters', 'daly', 'fleming', 'curtis', 'scott', 'wallace', 'alan', 'bennett', 'stephens', 'laurie', 'leonard', 'barnett', 'murphy', 'stuart', 'crawford', 'dunn', 'kemp', 'lester', 'ingram', 'connor', 'donald', 'mckay', 'bruce', 'hale', 'kirk', 'williamson', 'robinson', 'russell', 'barr', 'jim', 'abbott', 'donovan', 'morris', 'dickson', 'burke', 'chapman', 'keith', 'morrison', 'hartley', 'cameron', 'dennis', 'allen', 'bradshaw', 'thornton', 'gardner', 'townsend', 'evans', 'richards', 'steve', 'griffin', 'frank', 'atkinson', 'wills', 'donnell', 'doyle', 'moran', 'palmer', 'reynolds', 'bowen', 'bryan', 'slater', 'edwards', 'fisher', 'clarke', 'ramsey', 'brooks', 'cooper', 'gordon', 'harvey', 'morgan', 'ferguson', 'ross', 'chris', 'fred', 'smith', 'tom', 'david', 'cooke', 'benson', 'haynes', 'butler', 'cummings', 'matthews', 'perkins', 'hooper', 'taylor', 'brian', 'jenkins', 'buckley', 'hawkins', 'randall', 'michael', 'rogers', 'mcintyre', 'ted', 'phil', 'johnston', 'cullen', 'kelly', 'corbett', 'eric', 'clark', 'owen', 'rowe', 'connolly', 'moore', 'garrett', 'thompson', 'patrick', 'stephen', 'wade', 'brien', 'barrett', 'hart', 'saunders', 'james', 'nash', 'watkins', 'ellis', 'walsh', 'mason', 'todd', 'barnes', 'jennings', 'patterson', 'connell', 'lawson', 'craig', 'rodney', 'blake', 'adams']
\item Poincar\'e THR:
['nichols', 'mcintyre', 'randall', 'mckay', 'perkins', 'noel', 'elliott', 'reynolds', 'richards', 'glenn', 'lester', 'foley', 'walsh', 'murray', 'fitzgerald', 'donovan', 'riley', 'thornton', 'kemp', 'rodney', 'bartlett', 'kirk', 'bradley', 'curtis', 'jenkins', 'roberts', 'hayden', 'byron', 'skinner', 'smith', 'horton', 'carr', 'yates', 'chapman', 'benson', 'wilkinson', 'marshall', 'connor', 'bruce', 'barnett', 'quinn', 'fleming', 'barry', 'payne', 'carter', 'richardson', 'tanner', 'watson', 'freeman', 'buckley', 'simpson', 'watkins', 'owen', 'todd', 'miller', 'shaw', 'gibson', 'baker', 'ritchie', 'hooper', 'mason', 'osborne', 'lawson', 'harrington', 'jeremy', 'kerr', 'patterson', 'simmons', 'warren', 'wallace', 'jarvis', 'gardner', 'reilly', 'harvey', 'henderson', 'coleman', 'barrett', 'leonard', 'saunders', 'glover', 'hughes', 'farrell', 'anthony', 'fisher', 'cox', 'goodwin', 'bowman', 'mitchell', 'daniels', 'sullivan', 'griffin', 'abbott', 'morris', 'peterson', 'reeves', 'ralph', 'ross', 'elliot', 'brien', 'howard', 'donaldson', 'walters', 'russell', 'andrew', 'burke', 'edwards', 'dunn', 'phillips', 'hurley', 'lynch', 'rogers', 'barnes', 'doyle', 'harris', 'evans', 'stewart', 'stevenson', 'sheldon', 'burnett', 'connolly', 'burgess', 'cummings', 'williamson', 'wilson', 'steele', 'irwin', 'hicks', 'cooke', 'hanson', 'matthews', 'hawkins', 'gorman', 'willis', 'palmer', 'cameron', 'hayes', 'daly', 'morrison', 'moran', 'haynes', 'taylor', 'gordon', 'cunningham', 'stevens', 'dawson', 'clarke', 'morgan', 'robertson', 'mclean', 'thompson', 'spencer', 'murphy', 'davidson', 'duncan', 'evan', 'johnston', 'hart', 'terry', 'fletcher', 'spence', 'connell', 'griffith', 'parsons', 'allen', 'ellis', 'reid', 'adams', 'jones', 'nathan', 'norris', 'pearce', 'ingram', 'brooks', 'dillon', 'cooper', 'keith', 'crawford', 'hale', 'parker', 'webb', 'baxter', 'blake', 'turner', 'craig', 'fuller', 'nicholson', 'barker', 'campbell', 'fred', 'bailey', 'grady', 'nolan', 'welch', 'powell', 'armstrong', 'dalton', 'gavin', 'sanders', 'trevor', 'duffy', 'brent', 'dale', 'hoffman', 'garrett', 'boyd', 'robinson', 'dennis', 'jennings', 'clark', 'graham', 'kelley', 'newman', 'rowe', 'scott', 'phillip', 'porter', 'wright', 'ferguson', 'clayton', 'dixon', 'briggs', 'howell', 'mckenzie', 'chandler', 'collins', 'lewis', 'gallagher', 'mcbride', 'fowler', 'harding', 'flynn', 'lowe', 'moore', 'walker', 'bennett']
\item Graph:
['more', 'being', 'took', 'while', 'seen', 'under', 'never', 'are', 'by', 'furthermore', 'though', 'at', 'presumably', 'finally', 'then', 'ones', 'was', 'initially', 'these', 'their', 'among', 'together', 'or', 'also', 'included', 'few', 'up', 'earlier', 'even', 'existed', 'longer', 'first', 'be', 'beginning', 'hence', 'notably', 'the', 'well', 'will', 'similarly', 'actually', 'different', 'around', 'them', 'unlike', 'several', 'now', 'once', 'such', 'prior', 'fact', 'other', 'both', 'for', 'only', 'next', 'therefore', 'two', 'had', 'along', 'part', 'times', 'because', 'likewise', 'latter', 'since', 'additionally', 'to', 'where', 'still', 'than', 'but', 'end', 'taken', 'instance', 'addition', 'having', 'after', 'same', 'despite', 'of', 'similar', 'over', 'during', 'one', 'appear', 'outside', 'much', 'nevertheless', 'came', 'make', 'have', 'some', 'those', 'usually', 'it', 'this', 'number', 'when', 'separate', 'moreover', 'following', 'saw', 'with', 'time', 'before', 'full', 'within', 'perhaps', 'any', 'rest', 'might', 'others', 'exception', 'as', 'using', 'instances', 'is', 'making', 'found', 'made', 'use', 'come', 'without', 'until', 'should', 'example', 'through', 'so', 'itself', 'although', 'its', 'that', 'throughout', 'besides', 'they', 'consequently', 'ever', 'given', 'and', 'just', 'not', 'afterwards', 'there', 'added', 'years', 'on', 'later', 'however', 'could', 'ended', 'indeed', 'all', 'went', 'believed', 'take', 'rather', 'in', 'already', 'every', 'set', 'either', 'entered', 'possible', 'an', 'themselves', 'often', 'would', 'which', 'instead', 'second', 'last', 'each', 'thus', 'again', 'certain', 'most', 'old', 'from', 'were', 'yet', 'likely', 'elsewhere', 'been', 'way', 'new', 'what', 'own', 'has', 'out', 'if', 'another', 'many', 'particular', 'taking', 'can', 'today']
\end{itemize}

\end{document}